\documentclass{article}

\PassOptionsToPackage{numbers, sort&compress}{natbib}

\usepackage[final]{neurips_2023}

\usepackage[utf8]{inputenc} %
\usepackage[T1]{fontenc}    %
\usepackage{url}            %
\usepackage{booktabs}       %
\usepackage{amsfonts}       %
\usepackage{nicefrac}       %
\usepackage{microtype}      %
\usepackage{bbm}
\usepackage[dvipsnames]{xcolor}
\usepackage{tikz}
\usepackage[framemethod=tikz]{mdframed}
\usepackage{microtype}
\usepackage{graphicx}
\usepackage[caption=true,font=footnotesize, justification= centering]{subfig}
\usepackage{booktabs} %
\usepackage{enumitem}
\usepackage{amsmath}
\usepackage{amssymb}
\usepackage{amsthm}
\usepackage{xcolor}
\usepackage{tabularx}
\usepackage{makecell}

\usepackage{diagbox}
\usepackage{url}            %
\usepackage{booktabs}       %
\usepackage{amsfonts}       %
\usepackage{nicefrac}       %
\usepackage{microtype}      %
\usepackage{multirow}       %
\usepackage{algorithm}
\usepackage{algorithmic}

\usepackage{wrapfig}
\usepackage{comment}
\usepackage{amsmath,amssymb} %
\usepackage{symbols}
\usepackage{multirow}
\usepackage{ctable} %
\usepackage{pifont}%

\usepackage{amsmath,amsfonts,bm}

\def\1{\bm{1}}

\def\sp{space}

\def\rvh{{\mathbf{h}}}

\def\rvk{{\mathbf{k}}}

\def\rvx{{\mathbf{x}}}
\def\rvy{{\mathbf{y}}}

\def\rmH{{\mathbf{H}}}

\def\rmK{{\mathbf{K}}}

\def\rmX{{\mathbf{X}}}
\def\rmY{{\mathbf{Y}}}

\def\vy{{\bm{y}}}

\DeclareMathAlphabet{\mathsfit}{\encodingdefault}{\sfdefault}{m}{sl}
\SetMathAlphabet{\mathsfit}{bold}{\encodingdefault}{\sfdefault}{bx}{n}

\def\gD{{\mathcal{D}}}

\def\gF{{\mathcal{F}}}

\def\gL{{\mathcal{L}}}

\def\gR{{\mathcal{R}}}
\def\gS{{\mathcal{S}}}
\def\gT{{\mathcal{T}}}

\def\sR{{\mathbb{R}}}

\newcommand{\R}{\mathbb{R}}
\newcommand{\Co}{\mathbb{C}}

\newcommand*{\ShowNotes}{} %
\definecolor{darkred}{rgb}{0.7,0.1,0.1}
\definecolor{darkgreen}{rgb}{0.1,0.7,0.1}
\definecolor{cyan}{rgb}{0.7,0.0,0.7}
\definecolor{dblue}{rgb}{0.2,0.2,0.8}
\definecolor{maroon}{rgb}{0.76,.13,.28}
\definecolor{burntorange}{rgb}{0.81,.33,0}
\definecolor{tealblue}{rgb}{0.212,0.459, 0.533}

\definecolor{mypink}{rgb}{0.93359375, 0.62109375, 0.83984375}

\definecolor{pp}{rgb}{0.43921569, 0.18823529, 0.62745098}
\definecolor{rr}{rgb}{0.5254902 , 0.00784314, 0.12941176}
\definecolor{bb}{rgb}{0.09019608, 0.23529412, 0.37647059}
\definecolor{yy}{rgb}{0.49803922, 0.3372549 , 0.0}
\definecolor{gg}{rgb}{0.02352941, 0.3372549 , 0.17647059}

\ifdefined\ShowNotes
  \newcommand{\colornote}[3]{{\color{#1}\bf{#2: #3}\normalfont}}
\else
  \newcommand{\colornote}[3]{}
\fi

\usepackage[multiple]{footmisc}
\usepackage{wrapfig}

\definecolor{mybrown}{rgb}{0.87058824, 0.56078431, 0.01960784}
\definecolor{myblue}{rgb}{0.3372549 , 0.70588235, 0.91372549}
\definecolor{mypurple}{rgb}{0.8, 0.47058824, 0.7372549 }
\definecolor{myorange}{rgb}{0.835, 0.368, 0}
\definecolor{mygreen}{rgb}{0.00784314, 0.61960784, 0.45098039}
\definecolor{mygt}{rgb}{0.0078125 , 0.57421875, 0.40625}
\definecolor{mysp}{rgb}{0.84765625, 0.515625  , 0.0234375}
\definecolor{mycitecolor}{rgb}{0,0.08,0.45}
\definecolor{mygr}{rgb}{0.9607,0.9607,0.9607}
\definecolor{myoo}{rgb}{0.992,0.9176,0.9019}

\definecolor{myrr}{HTML}{AE031A}
\definecolor{mybb}{HTML}{0155B3}

\usepackage{pifont}
\usepackage[T1]{fontenc}
\usepackage[utf8]{inputenc}
\usepackage{pmboxdraw}
\usepackage[font={footnotesize}, labelfont=bf,labelsep=period]{caption}
\usepackage{thm-restate}
\definecolor{cvprblue}{rgb}{0.21,0.49,0.74}
\definecolor{lightcarminepink}{rgb}{0.9, 0.4, 0.38}
\usepackage[pagebackref,breaklinks,colorlinks,citecolor=cvprblue,linkcolor=lightcarminepink]{hyperref}

\mdfdefinestyle{MyFrame}{%
    linecolor=Black,
    outerlinewidth=0.1pt,
    roundcorner=2pt,
    innertopmargin=0.3\baselineskip,
    innerbottommargin=0.3\baselineskip,
    innermargin = 0.2cm,
    outermargin = 0.cm,
    innerrightmargin=5pt,
    innerleftmargin=5pt,
    backgroundcolor=mybb!3!white}
\usepackage[multiple]{footmisc}

\usepackage[textsize=tiny]{todonotes}

\usepackage{listings}

\title{Truly Scale-Equivariant Deep Nets\\ with Fourier Layers}

\author{%
Md Ashiqur Rahman \quad Raymond A. Yeh\\
  Department of Computer Science, Purdue University\\
  \texttt{\{rahman79, rayyeh\}@purdue.edu} \\
}

\begin{document}
\maketitle
\begin{abstract}
In computer vision, models must be able to adapt to changes in image resolution to effectively carry out tasks such as image segmentation; This is known as scale-equivariance. Recent works have made progress in developing scale-equivariant convolutional neural networks, e.g., through weight-sharing and kernel resizing. However, these networks are not truly scale-equivariant in practice. Specifically, they do not consider anti-aliasing as they formulate the down-scaling operation in the continuous domain. To address this shortcoming, we directly formulate down-scaling in the discrete domain with consideration of anti-aliasing. We then propose a novel architecture based on Fourier layers to achieve truly scale-equivariant deep nets, i.e., absolute zero equivariance-error. Following prior works, we test this model on MNIST-scale and STL-10 datasets. Our proposed model achieves competitive classification performance while maintaining zero equivariance-error. The code is available at \url{https://github.com/ashiq24/Scale_Equivarinat_Fourier_Layer}.
\end{abstract}

\section{Introduction}

\begin{figure*}[t]
    \centering
    \subfloat[Illustration that regular CNNs are not scale-equivariant.]{
    \hspace{-0.3cm}
    \includegraphics[width=.5\linewidth]{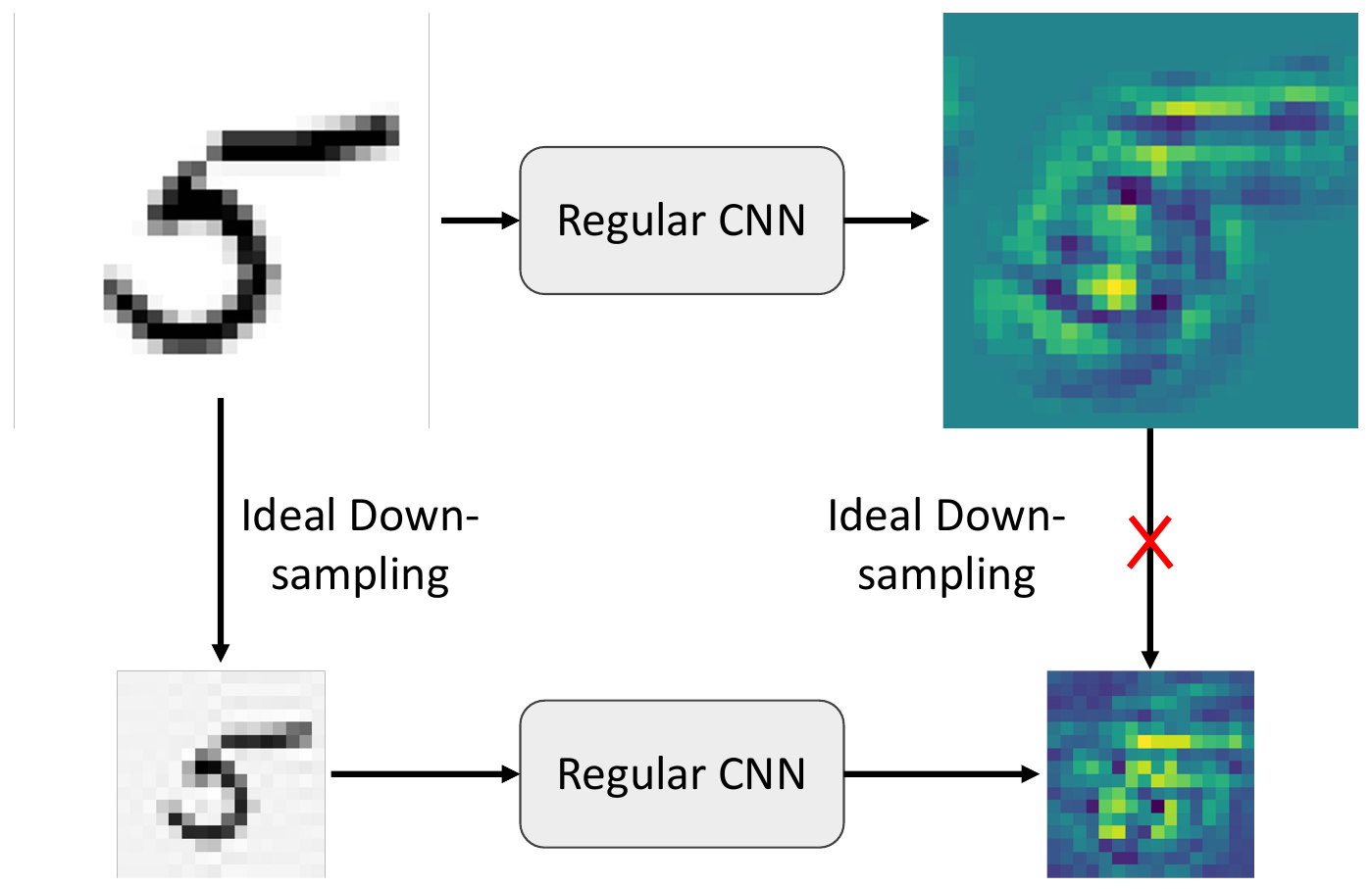}
    \label{fig:scale_eq_base}\hspace{0.02\linewidth}
    }
    \subfloat[Illustration that our model is scale-equivaraint.]{
    \hspace{-0.35cm}
    \includegraphics[width=.49\linewidth]{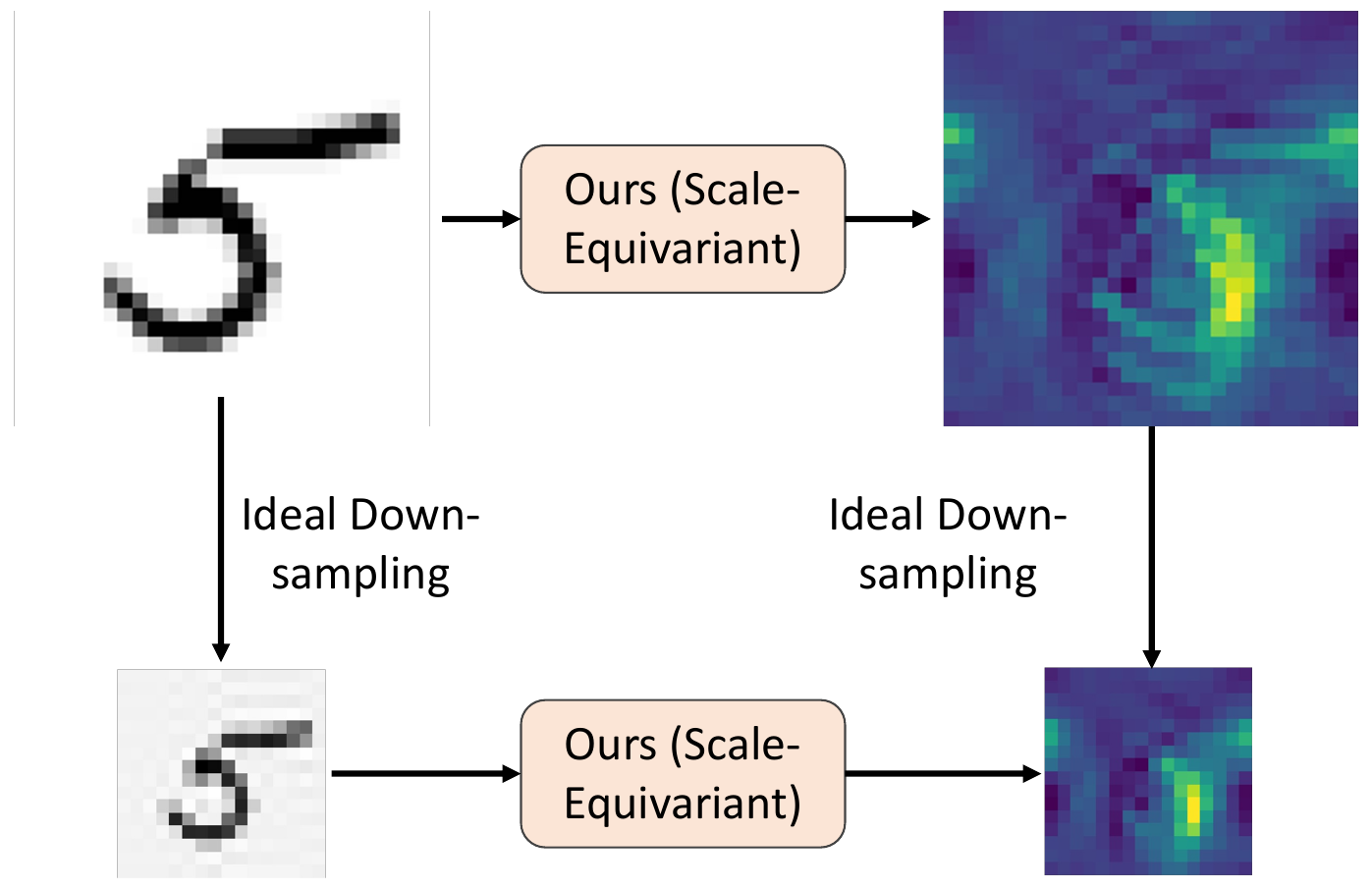}
    \label{fig:scale_eq_ours}
    }
    \caption{Comparison of scale-equivariance on CNN vs. our model. For a regular CNN, the features extracted from the corresponding high/low image resolution look very different. On the other hand, downsampling the high-res feature is guaranteed to achieve the same feature obtained from the low-resolution image.
    }
    \label{fig:ill_eq}
\end{figure*}

Consider the task of image classification; if an object in the image is scaled (resized), then its corresponding object label should remain the same, \ie, scale-invariant. Similarly, for semantic segmentation, if an object is scaled, then its corresponding mask should also be scaled accordingly, \ie, scale-equivariant. Similarly, one would expect the features extracted to be scale-equivariant; see~\figref {fig:ill_eq} for illustration.
These invariant and equivariant properties are important to many computer vision tasks due to the nature of images. A photo of the same scenery can be taken from different distances, and objects in the scenes may come in different sizes. Developing representations that effectively capture this multi-resolution aspect of images has been a long-standing quest~\cite{adelson1984pyramid,grauman_2005_pyramid, lazebnik2006beyond,he2015spatial,zhao2017pyramid}.

Recently, there has been a line of work on developing scale-equivariant convolutions networks~\cite{kanazawa2014locally,ghosh2019scale,worrall2019deep,sosnovik2019scale,sosnovik2021disco} to more effectively learn multi-resolution features. At a high level, these works achieve scale-equivariant convolution layers through weight-sharing and kernel resizing, \ie, use the ``same'' but resized kernel across all scales~\cite{cohen2016group}. The innovation of these works is how to properly resize the kernel. For example,~\citet{bekkers2019b} and~\citet{sosnovik2019scale} formulate kernel resizing as a continuous operation and then discretize the kernel when implemented in practice. However, this discretization leads to non-negligible equivariance error. On the other hand,~\citet{worrall2019deep} and ~\citet{sosnovik2021disco} directly formulate kernel resizing in the discrete domain, \eg, using dilation or solving for the best kernel given a fixed scale set, and achieve low equivariance-error. 

Despite these successes, we point out that the aforementioned works are not truly scale-equivariant in practice. Specifically, these works are derived using a continuous domain down-scaling operation, \ie, there is no need to consider anti-aliasing. However, when performing a down-scaling on discrete space, the Nyquist theorem~\cite{nyquist1928certain,manolakis2011applied} tells us that an anti-alias filter is necessary to avoid high-frequency content to alias into lower frequencies. The canonical example of aliasing is the ``wagon-wheel effect'', where a wheel in a video appears to be rotating slower or even in reverse from its true rotation. To address this gap from prior work, we consider the down-scaling operation directly in the discrete domain, taking he anti-aliasing into account.

In this work, we formulate down-scaling as the ideal downsampling from signal processing~\cite{manolakis2011applied}. We then propose a family of deep nets that are truly scale-equivariant based on this ideal downsampling. 
This involves rethinking all the components in the deep net, including convolution layers, non-linearities, and pooling layers. 

With the developed deep net, we focus on the task of image classification. We further point out that truly scale-invariant classifiers are not desirable. A truly scale-invariant model's performance is limited by the lowest-resolution image. Instead, the more desirable property is that a high-resolution image should achieve a better performance than its corresponding low-resolution image. This motivated us to design a classifier architecture suitable for this property.

Following prior works, we conduct our experiments on the MNIST-scale~\cite{sohn2012learning} and STL~\cite{coates2011analysis} dataset. By design, our method achieves zeros scale equivariance-error both in theory and in practice. In terms of accuracy, we compare to recent scale-equivariant CNNs. We found our approach to be competitive in classification accuracy and exhibit better data efficiency in low-resource settings.

{\bf \noindent Our contributions are as follows:}
\begin{itemize}[leftmargin=0.45cm]
\vspace{-0.1cm}
\itemsep0.25em 
    \item We formulate down-scaling in the discrete domain with considerations of anti-aliasing.
    \item We propose a family of deep nets that is truly scale-equivariant by designing novel scale-equivariant modules based on Fourier layers.
    \item We conduct extensive experiments validating the proposed approach. On MNIST and STL datasets, the proposed model achieves an absolute zero end-to-end scale-equivariance error while maintaining competitive classification accuracy.
\end{itemize}

\section{Related Work}

{\noindent \bf Scale-equivariance and invariance.}
The notation of scale-equivariance is deeply rooted in image processing and computer vision. For example, 
classic hand-designed scale-invariant features such as SIFT~\cite{lowe1999object,lowe2004distinctive} have made tremendous contributions to the field of computer vision. Earlier works propose to use an image or spatial pyramid to capture the multi-resolution aspect of an image~\cite{adelson1984pyramid,grauman_2005_pyramid, lazebnik2006beyond} by extracting features at several scales in an efficient manner.

More recently, there have been interests in developing scale-equivariant CNN~\cite{kanazawa2014locally,ghosh2019scale,bekkers2019b,worrall2019deep,sosnovik2021disco,sosnovik2019scale,zhu2022scaling}. Based on Group-Conv~\cite{cohen2016group}, these works achieve scale-equivariance convolution layers through weight-sharing and kernel resizing. Different from these works, we consider the down-scaling in the discrete domain formulated as ideal downsampling from signal processing. We then develop modules that are truly scale-equivariant to enable a deep net that achieves zero equivariance-error measured from end to end.
Finally, we note that there is a rich literature of equivariant deep nets~\cite{cohen2016group,bronstein2017geometric,ravanbakhsh2017equivariance,weiler2019general,venkataraman2019building,romero2020attentive,yeh2022equivariance,shakerinava21a,rojas2023making} with numerous applications applied to various domains, \eg, sets~\cite{ravanbakhsh_sets,zaheer2017deep, qi2017pointnet, hartford2018deep,yeh2019chirality,maron2020learning}, graphs~\cite{shuman2013emerging,defferrard2016convolutional,kipf2017semi,maron2018invariant,yeh2019diverse,dehaan2020gauge,liu2020pic,liu2021semantic,morris22a}, \etc.  Moreover, several recent studies have also identified and tackled the issue of aliasing generated from the pooling layer to attain finer translation equivariance~\cite{zhang2019making,xu2021group,rojas2022learnable} and image generation \cite{karras2021alias}.

{\noindent \bf Fourier transform in neural networks.}
Fourier transforms have been previously used in deep learning. For example,~\citet{mathieu2013fast} proposes to use Fast Fourier Transform (FFT) to speed up CNN training. Fourier transform has also been used to develop network architectures, including various convolutional neural networks operating in the Fourier space~\cite{pratt2017fcnn,kondor2018clebsch}. Recently, Fourier layers are capable of handling inputs of varying resolution, which have been employed in neural operators, facilitating applications in partial differential equations and in state space models  ~\cite{nguyen2022s4nd,li2021fourier}. Fourier convolutions have also found success in low-level image processing tasks, \eg, inpainting~\cite{suvorov2022resolution}, deblurring~\cite{xint2023freqsel}. Different from these works, we focus on developing truly scale-equivariant deep nets and leverage Fourier layers to achieve this goal.

\section{Preliminaries}
We briefly introduce and review the definition of Fourier transform, ideal downsampling, and scale-equivariance. For readability, we use 1D data to define these concepts. These ideas are extended to 2D data with multiple channels when implemented in practice.

{\bf\noindent Discrete Fourier Transform (DFT).}
Given an input vector $\rvx \in \R^N$, we consider $\gF: \R^N \rightarrow \Co^N$ be the discrete Fourier Transform (DFT) which has the form  
\be\label{eqn:fft}
    \rmX = \gF(\rvx)  \text{ such that }~\rmX[k] \triangleq \frac{1}{N} \sum_{n = 0}^{N-1} \rvx[n] e^{-j \frac{2\pi}{N}kn}, 
\ee
where $j$ denotes the unit imaginary number, \ie, $j^2=-1$. The index $k$ in~\equref{eqn:fft} is commonly within the domain of $[0,N]$. Note that as~\equref{eqn:fft} is $N$-periodic, for readability, we will use $k$ from $[-\frac{N-1}{2}, \frac{N-1}{2}]$ where $k=0$ corresponds the lowest frequency. 

The corresponding inverse DFT (IDFT) $\gF^{-1}: \Co^N \rightarrow \R^N$ is defined as 
\be\label{eqn:ifft}
    \rvx = \gF^{-1}(\rmX) \text{ such that }~\rvx[n] = \sum_{k = 0}^{N-1} \rmX[k] e^{j \frac{2\pi }{N}kn}.
\ee

By the convolution property of DFT, the circular convolution between $\rvx$ and a kernel $\rvk \in \R^N$ can be represented as the element-wise multiplication in the Fourier domain,~\ie,
\bea
\gF(\rvx \circledast \rvk) = \gF(\rvx) \odot \gF(\rvk) = \rmX \odot \rmK,
\eea
where $\circledast$ denotes the circular convolution and $\odot$ denotes element-wise multiplication. Unless explicitly mentioned, we will represent the input vector with lowercase letters (e.g., $\rvx$) and its corresponding DFT with uppercase letters (e.g., $\rmX$).

\begin{figure*}[t]
    \centering
    \subfloat[Illustration of ideal anti-aliasing (low-pass) filter.]{
    \hspace{-2cm}
    \includegraphics[height=3.3cm]{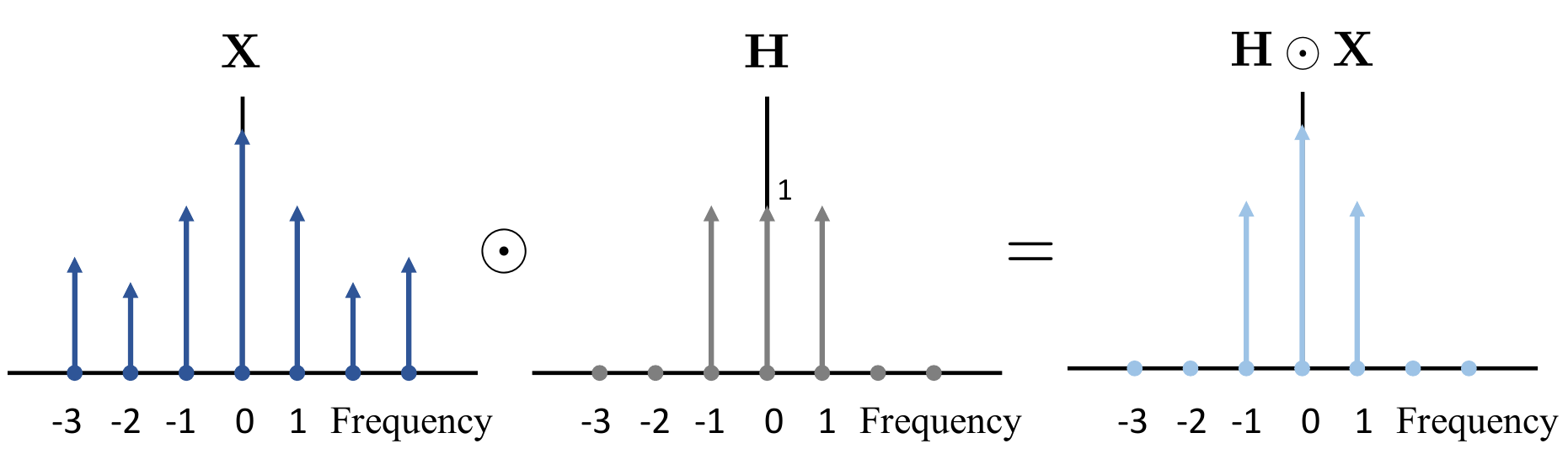}
    \label{fig:lfp}\hspace{0.02\linewidth}
    }
    \subfloat[Illustration of Claim~\ref{cla:main_claim}.]{
    \includegraphics[height=3.35cm]{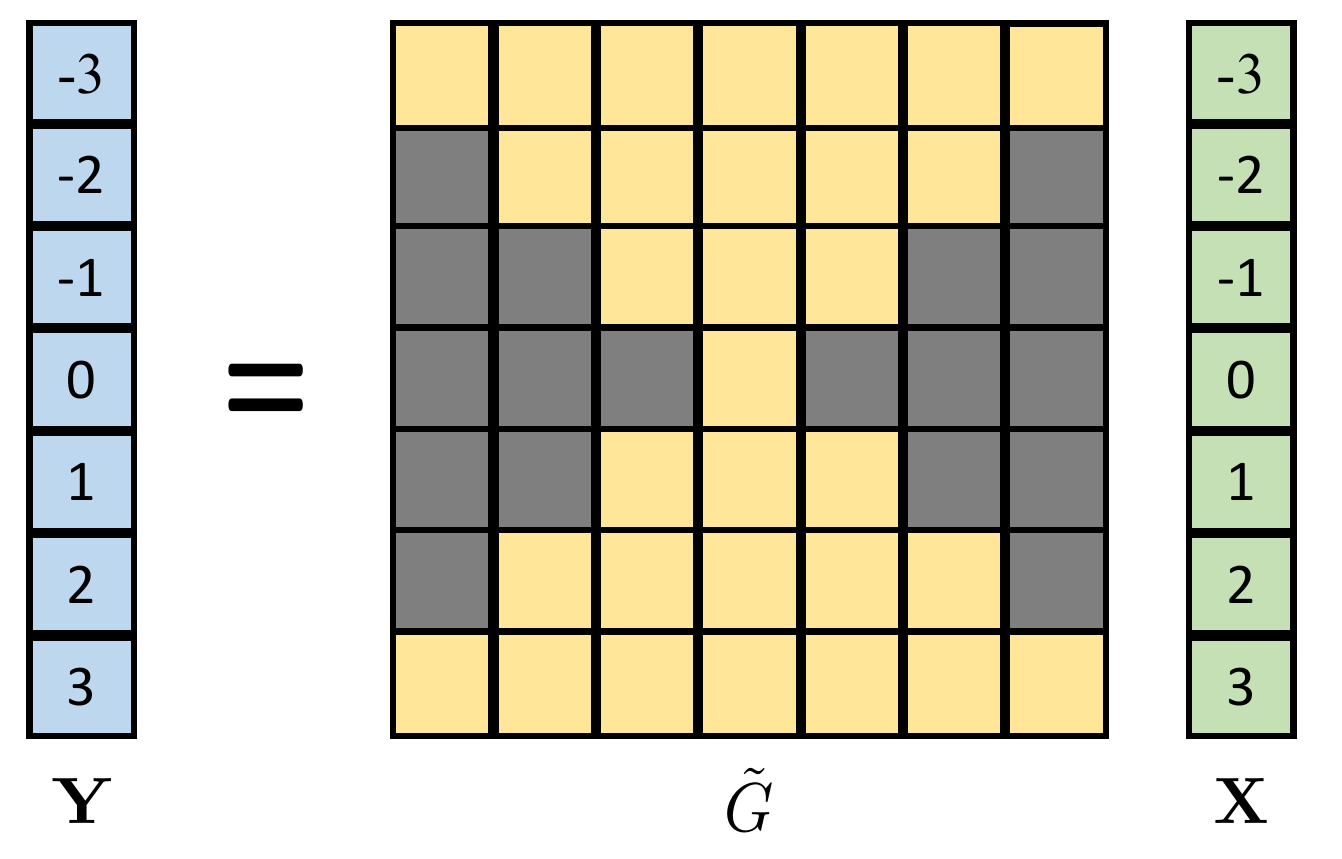}
    \label{fig:eq_app}
    }
    \caption{In (a), we illustrate an ideal low-pass filter showing that it zeros out the high frequencies. 
    In (b), we illustrate the structure described in Claim~\ref{cla:main_claim} for a linear $\tilde{G}$. The {\color{darkgray} \bf gray regions} correspond to the value being zero.
    }
\end{figure*}

{\bf\noindent Down-scaling operation.} To reduce the scale (or resolution) of a signal $\rvx \in \sR^N$, one could perform a subsampling ${\tt Sub}_R$ by a factor of $R$
\bea
{\tt Sub}_R(\rvx)[n] = \rvx[Rn].
\eea
However, naively subsampling leads to aliasing. Hence, anti-aliasing is performed in a multi-rate system. In signal processing, the analysis commonly uses the ideal anti-aliasing
filter $\rvh$, which zeros out all the high-frequency content, \ie, its DFT $\rmH \triangleq \gF(\rvh)$ is defined as:
\bea
\rmH[k] = 1~\text{if }~|k| \leq \frac{N}{2R}~\text{and}~ 0~\text{otherwise}.
\eea
See~\figref{fig:lfp} for an illustration of the ideal anti-aliasing filter.

In this work, we define the overall down-scaling operation to be the ideal downsampling $\gD_R$ by a factor of $R$, which performs anti-aliasing followed by a subsampling operation: 
\bea\label{eq:scale_opt}
\gD_R(\rvx) \triangleq {\tt Sub}_R(\rvh \circledast \rvx)~~\forall R<N,
\eea
where their DFT are related by
\bea\label{eq:scale_opt_fourier_rel}
\gF(\gD_R(\rvx)) = \gF(\rvx)\left[-N/2R:N/2R\right].
\eea
{\bf\noindent Scale-equivariance.} With the down-scaling operation defined, a deep net ${\tt g}: \{\sR^{1}, \sR^{2}, \dots, \sR^{N} \} \mapsto \{\sR^{1}, \sR^{2}, \dots, \sR^{N} \}$ is scale-equivariant if:
\bea\label{eq:scale_eq}
{\tt g}(\gD_R(\rvx)) = \gD_R({\tt g}(\rvx)) \quad\forall \rvx \in \{\sR^{1}, \sR^{2}, \dots, \sR^{N} \} \text{ and } R < \text{dim}(\rvx),
\eea
where $\{\sR^{1}, \sR^{2}, \dots, \sR^{N} \}$ represents the space of input/output signals at different scales.
In this paper, we are interested in designing a family of deep nets that satisfies the equality in~\equref{eq:scale_eq}. Scale-invariance can be defined in a similar manner as
\bea\label{eq:scale_invar}
{\tt g}(\gD_R(\rvx)) = {\tt g}(\rvx) \quad\forall \rvx \in \{\sR^{1}, \sR^{2}, \dots, \sR^{N} \} \text{ and } R < \text{dim}(\rvx).
\eea

{\noindent \bf Fourier layer.} 
Given a multi-channel input vector, $\rvx \in \R^{C_{\tt in} \times N}$ and kernel $\rvk \in \R^{C_{\tt out} \times C_{\tt in} \times N}$, where $C_{\tt in/out}$ is the number of input/output channels, the circular convolution layer is defined as
\begin{equation}\label{eq:fourier_layer}
    \gF((\rvx \circledast \rvk))[c'] = \sum_{c=1}^{C_{\tt in}} \rmX[c] \odot \rmK[c',c],
\end{equation}
where $\rmX$ and $\rmK$ denotes the DFT of $\rvx$ and $\rvk$ applied independently for each channel.

\section{Approach}
Our goal is to design truly scale-equivariant deep nets. To accomplish this goal, we propose scale-equivariant versions of CNN modules, including, the convolution layer, non-linearities, and pooling layers. In~\secref{sec:eq_net}, we detail the operation for each of the proposed modules.
In~\secref{sec:beyond}, we demonstrate how to build a classifier that is suitable for image classification with scale-equivariant features. We now explain our overarching design principle for the scale-equivariant modules. 

From a frequency perspective, as reviewed in~\equref{eq:scale_opt}, the ideal downsampling operation results in the loss of higher frequency terms of the signal. In other words, if a feature's frequency terms depend on any higher frequency terms of the input, then it is not scale-equivariant, as the information will be lost after downsampling. We now formally state this observation in Claim~\ref{cla:main_claim}. 

\begin{mdframed}[style=MyFrame,align=center]
\begin{claim}\label{claim:all}
\vspace{.2em}
Let $g$ denote a deep net such that $\rvy = g(\rvx)$.
If this deep net $g$ can be equivalently represented as a set of functions $\tilde{G}_k: \Co^{2k+1} \rightarrow \Co$ such that
\bea
\rmY[k]= \tilde{G}_k (\rmX[-k:k]
)~~\forall k 
\eea
then $g$ is scale-equivariant as defined in~\equref{eq:scale_eq}. In other words, an output's frequency terms can only have dependencies on the terms in $\rmX$ that are {\bf even lower} in frequencies. We illustrate this structure with a linear function in~\figref{fig:eq_app}.
\label{cla:main_claim}
\end{claim}
\end{mdframed}

\begin{proof}
\vspace{-0.35cm}
We denote the deep net's input and output as $\rvx$ and $\rvy$ with corresponding DFT $\rmX$ and $\rmY$.
We denote the deep net's down-scaled input and output as $\rvx' = \gD_R(\rvx)$ and $\rvy'=g(\rvx')$ with corresponding DFT $\rmX'$ and $\rmY'$. 
Now assume that $g: \R^n \rightarrow \R^n ~~ \forall n \in \{1,2, \dots N\}$ is a deep net that satisfies Claim~\ref{cla:main_claim} then
\bea
&\rmY[k] &= \tilde{G}_k(\rmX[-k:k]) ~~ \forall k \le \frac{N}{R}\\
&         &=\tilde{G}_k(\rmX'[-k:k]) = \rmY'[k] \text{~~Following the property of } \gD_R \text{ in~\equref{eq:scale_opt_fourier_rel}} 
\eea
Therefore, $\forall k \le \frac{N}{R}~~ \rmY[k] = \rmY'[k]$. By the definition of ideal downsampling $\rmY' = \gD_R(\rmY)$ concluding that ${g}(\gD_R(\rvx)) = \gD_R({g}(\rvx))$, \ie, $g$ is scale-equivariant.

For ease of understanding, here we assume that the deep net's input and output are of the same size. A version with a more relaxed assumption is provided in Appendix~\secref{supp_sec:claim}. 
\end{proof}

\subsection{Scale Equivariant Fourier Networks.}\label{sec:eq_net}%
We now describe the proposed modules and show that they are truly scale-equivariant.

{\bf\noindent Spatially local Fourier layer.}
For computer vision, learning local features is crucial. The Fourier layer in~\equref{eq:fourier_layer} is global in nature. To efficiently learn local features, we propose a localized Fourier layer where we constrain the degree of freedom in the kernel $\rmK$ such that the respective spatial kernel $\rvk$ is spatially localized. 

Let $\rvk \in \R^d$ and $\rvk^l \in \R^l$ to be $d$ and $l$ dimensional kernel such that 
 $\rvk[i] = \rvk^l[i] ~\text{if } i<l \text{ otherwise } 0$, \ie, $\rvk$ is spatially local to have a receptive field of size $l$.
 We denote $\rmK$ and $\rmK^l$ be the DFT of the kernel $\rvk$ and $\rvk^l$ respectively.
 We claim that $\rmK$ can be written as 
\bea\label{eq:local_kernel}
 \rmK[p] = \frac{1}{d} \sum_{m = -\frac{l}{2}}^{\frac{l}{2}} \big( \rmK^l[m] \sum_{n=0}^{l-1}e^{-2 \pi jn( \frac{p}{d}- \frac{m}{l})} \big).
\eea
From~\equref{eq:local_kernel}, instead of modeling all the degrees of freedom in $\rmK$, we will directly parameterize $\rmK^l$ to enforce the learned kernel to be localized spatially. We defer the proof to the Appendix~\secref{supp_sec:proof}.

\vspace{0.1cm}
\begin{mdframed}[style=MyFrame,align=center]
\begin{claim}
\vspace{0.2em}
The spatially local Fourier layer is scale-equivariant.
\end{claim}
\end{mdframed}
\begin{proof}\vspace{-0.4cm}
The kernel $\rvk$ has a corresponding DFT $\rmK$. As reviewed, a circular convolution between $\rvk$ and input $\rvx$ can be expressed as 
\bea
\rmY[k] = \rmK[k] \odot \rmX[k]~\forall k.
\eea
Observe that $\rmX[k]$ is a subset of $\rmX[-k:k]$, \ie, Claim~\ref{cla:main_claim} is satisfied.
\vspace{-0.1cm}
\end{proof}

{\bf\noindent Scale-equivariant non-linearity ($\sigma_{\tt s}$).}
Element-wise non-linearities, \eg, ReLU, in the spatial domain are generally not scale-equivariant under the ideal downsampling operation ${\gD}_R$. While applying element-wise non-linearity in the frequency domain is scale-equivariant, this strategy empirically leads to degraded performance on classification tasks. To address this, we propose a scale-equivariant non-linearity $\sigma_{\tt s}$ in the spatial domain. 

Given a non-linearity $\sigma$,~\eg, ReLU, we construct a corresponding scale-equivariant version $\sigma_{\tt s}$ that satisfies Claim~\ref{cla:main_claim}. Let $\rvx \in \R^N$ and $\rvy \in \R^N$ to denote the input and output of $\sigma_{\tt s}$. We define scale-equivariant non-linearity $\sigma_{\tt s}(\rvx) = \gF^{-1}(\rmY)$ where $\rmY$ takes the following form:
\begin{equation}
 \rmY[k] = \left\{
                \begin{array}{ll}
                \gF \Big( \sigma \circ \gF^{-1}(\rmX {\color{mybb}\big[}0{\color{mybb}\big]}) \Big)[0], &  k = 0 \\
                \vspace{-0.17cm}&\\
                \gF\Big( \sigma \circ \gF^{-1}(\rmX{\color{mybb}\big[}-1:1{\color{mybb}\big]}) \Big)[1],   & k = 1 \\
                \vdots & \vdots \\
               \gF\Big( \sigma\circ \gF^{-1} (\rmX{\color{mybb}\big[}-|k|:|k|{\color{mybb}\big]} ) \Big)[k],  & k = k \\
                \vdots & \vdots \\
                \end{array} 
\right.
\end{equation}
$\forall |k| \le \frac{N}{2}$ and $\rmX$ denotes the DFT of the input, \ie, $\gF(\rvx)$. In practice, we choose $\sigma$ to be ReLU in our implementation.

Next, it is generally computationally expensive to achieve equivariance over all scales. In practice,
we only enforce a set of scales for which we want to achieve equivariance, which can be denoted in terms of corresponding resolutions as $\gR = (m, \dots N)$ with $\gR[i] < \gR[i+1]$. To achieve scale-equivariant non-linearity over the scales of $\gR$, $\sigma_{\tt s} = \gF^{-1}(\rmY)$ can be efficiently computed as

\bea\label{eq:fast}
\rmY[k] =  \gF\Big( \sigma \circ \gF^{-1}(\rmX{\color{mybb}\big[}-\frac{\gR'[i]}{2}:\frac{\gR'[i]}{2}{\color{mybb}\big]}) \Big)[k]~~\text{~for the}~i~\text{s.t.~} \frac{\gR'[i-1]}{2} < |k| \le \frac{\gR'[i]}{2}.
\eea
Here, the ordered set $\gR' = \gR \cup \{0\}$. By~\equref{eq:fast}, all the Fourier coefficients $k$ between any two consecutive resolutions in $\gR$, \ie, $\gR[i-1]/2<|k| \le \gR[i]/2$ can be computed by a single Fourier transform pair.

{\bf\noindent Scale-equivariant pooling.}
Pooling operation is crucial for deep nets' scalability to larger images and datasets as they make the network more memory and computationally efficient. Commonly used pooling operations are max/average pooling, which reduces the input size by the factor of its window size $w$ and is not scale-equivariant. To address this, we propose scale-equivariant pooling $\texttt{Pool}^w_{\tt s}$.

Let ${\tt Pool}^{w}$ denote a max/average pooling operation with a window size $w$ and ${\tt Pool}^{w}: \R^d \rightarrow \R^{\frac{d}{w}}$. 

We define scale-equivariant pooling operation ${\tt Pool}^{w}_{\tt s}: \R^d \rightarrow \R^{\frac{d}{w}}$ mapping from $\rvx$ to $\rvy$ where $\rvy = \gF^{-1}(\rmY)$ follows
\bea
\rmY[k] = \gF\big( {\tt Pool}^{\tt w}( \gF^{-1} (\rmX{\color{mybb}\big[}-w |k|: w |k|{\color{mybb}\big]} ) ) \big)[k] ~~~~~ \forall k \le \frac{d}{2 w}.
\eea
Observe that this pooling layer satisfies Claim~\ref{cla:main_claim} by construction. Similar to non-linearity, we can enforce the equivariance over the set $\gR$ following the same formulation in~\equref{eq:fast}.

Note that as pooling reduces the size of the output by a factor $w$, the operation is only scale-equivariant at every $w^{\text{th}}$ resolution. When the input size is not a multiple of $w$ there is a truncation of the input.

{\bf\noindent Time Complexity.}
We now provide the time complexity of our scale-equivariant Fourier layer and compare it with standard group convolutions. Let's consider a 1D signal of length $N$ and a kernel of length $K$. Our proposed model involves:
\begin{itemize}
    \item A transformation of local filter to global with time complexity $O(KN)$
    \item A convolution using Fourier transform with time complexity $O(N log (N))$
    \item Our scale equivariant non-linearity depends on the size of the group. Let $A$ be the set of group actions. The time complexity of the proposed scale-equivariant non-linearity is $O(|A| N \log (N))$, where $|A|$ denotes the cardinality of the set $A$.
\end{itemize}
So, the time complexity for each layer becomes $$O(|A| N \log (N) + KN).$$ 

As a comparison, the time complexity of regular group convolution is $O(KN|A|)$ in the first layer and $O(KN|A|^2)$ for all intermediate layers, assuming the cost of group action is a negligible constant \cite{he2021efficient}. 

Considering the time complexity of the intermediate layers of group convolutions, our proposed method is more efficient when 
$$|A| N \log (N) + KN < |A|^2KN \implies \log (N) + \frac{K}{|A|} < |A|K.$$

So, when $K << |A|$ and $ log (N) < |A|K$, i.e.,  assuming the set of group actions of moderate size, then our method is faster than group convolutions. 

Modern GPUs are specifically optimized for regular convolution operations that can be performed in place. In contrast, the FFT algorithm does not fully capitalize on GPUs' advantages, primarily due to unique memory access patterns and moderate arithmetic intensities. Consequently, our approach is unable to harness the full potential of GPUs. When executed on a GPU, regular group convolutions implemented as standard convolutions might exhibit comparable or even shorter running times than our approach.

\subsection{Classifier for equivariant features}\label{sec:beyond}
A truly scale-invariant, defined in~\equref{eq:scale_invar}, the model's performance is limited by the lowest resolution as the prediction needs to be the same. In the extreme, the prediction can only depend on a single mean pixel. Instead of invariance, we believe that it is more desirable to ensure that a high-resolution image achieves a better performance than its down-scaled version,~\ie, the performance is scale ``consistent''. To achieve this property, we propose a suitable classifier architecture and training scheme.

{\noindent\bf Classifier.} In order to enforce scale-consistency, we need a classifier that outputs a prediction per scale. This motivated the following proposed architecture.
Let $c$ be a classifier with $M$ classes where $\hat{\vy} = {c} \circ {g} (\rvx) \in \R^{|\gR(\rvx)| \times M}$. $\gR(\rvx)$ is defined as the set of resolutions smaller than the input resolution in the considered scales $\gR$.~\ie, $\gR(\rvx) = \{k: k\le {\text{dim}}(\rvx) \text{ and } k \in \gR\}$. Here, $g$ is a scale-equivariant deep-net that extracts features $\phi=g(\rvx)$ with corresponding DFT of $\Phi$.
Our proposed classifier has the form: 
\bea 
\hat{\vy}[k] = {\tt MLP} \circ {\tt Pool} \bigg(\gF^{-1} \big( {\tt{Pad}}_{\tt N} (\Phi{\color{mybb}[} -\frac{|k|}{2} : \frac{|k|}{2}{\color{mybb}]})\big) \bigg) ~~ \forall k \in \gR(x)
\eea
where $\tt Pad_N$ is a Fourier padding operation that symmetrically pads zero to either side of the DFT to a fixed size $\tt N$, $\tt Pool$ is a spatial pooling operation and $\tt MLP$ maps the pooled feature to the predicted logits $\hat\vy[k]$ for each scale; Note the $\tt MLP$ is shared across all scales. As we are sharing the MLP, we need to ensure that the input sizes are identical. Hence, we padded the features $\Phi$ to a fixed size. Finally, at test-time,  we use the output from $\hat{\vy}[{\text{dim}(\rvx)}]$ to make a prediction.

{\noindent\bf Training.} 
Given a dataset $\gT = \{(\rvx, y)\}$, we train our model using the sum of two losses. The first term is a standard sum of cross entropy loss $\gL$ over the scales:
\bea
\sum_{k \in \gR(\rvx)} \gL(\hat{\vy}[k],y).
\eea
The second term is a consistency loss to encourage the performance of high-resolution to be better than the low-resolution:
\bea\label{eq:hindge}
\sum_{k \in \gR(\rvx)} {\max} \bigg(\gL(\hat{\vy}[k],y) -\gL(\hat{\vy}[k-1],y), 0\bigg).
\eea

This is a hinge loss that penalizes the model when the cross entropy loss $\gL$ on high-resolution features (larger $k$) is greater than that of the low-resolution features (smaller $k$).

\section{Experiments}
To study the effectiveness of our model, we conduct experiments on two benchmark datasets, MNIST-scale~\cite{sohn2012learning} and STL10~\cite{coates2011analysis}, following our theoretical setup using ideal downsampling. In this case, the theory exactly matches practice, and our approach achieves perfect scale-equivariance. We also conduct experiments comparing the models' generalization to unseen scales and data efficiency. Finally, we conduct experiments using a non-ideal anti-aliasing filter in down-scaling. Under this setting, our model no longer achieves zero scale equivariance-error. However, we are interested in how the models behave under this mismatch in theory and practice.

\begin{table}[t]
\begin{minipage}[c]{0.49\textwidth}
\centering
\small
\caption{Accuracy of different models on MNIST-scale (ideal downsampling) with all scales.
} 
\vspace{0.06cm}
\begin{tabular}{@{}lccc}
\specialrule{.15em}{.05em}{.05em}
Models                                     & Acc.$\uparrow$           & Scale-Con.$\uparrow$ & Equi-Err.$\downarrow$   \\ \midrule
\multirow{1}{*}{CNN}                       & 0.9737          & 0.6621       &  -           \\
\multirow{1}{*}{Per Res. CNN}              & 0.9388          &  0.0527      &   -           \\
\hline    
\multirow{1}{*}{SESN}                      & 0.9791          & \underline{0.6640}       &     -          \\  
\multirow{1}{*}{DSS}                       & 0.9731         &  0.6503       &    -         \\    
\multirow{1}{*}{SI-CovNet}                 &0.9797          & 0.6425       &    -         \\  
\multirow{1}{*}{SS-CNN}                    &0.9613          & 0.3105        &   -           \\
\multirow{1}{*}{DISCO}                     & \underline{0.9856}        & 0.5585         &0.44          \\  \midrule
\multirow{1}{*}{Fourier CNN}               & 0.9713        & 0.2421         & \underline{0.28}         \\
\multirow{1}{*}{Ours}                      & \textbf{0.9889} & \textbf{0.9716}      &   \textbf{0.00}      \\
\specialrule{.15em}{.05em}{.05em}
\end{tabular}
\label{tab:mnist_result_ideal}

\end{minipage}
\hspace{0.15cm}
\begin{minipage}[c]{0.49\textwidth}
\centering
\small
\caption{
Accuracy of different models on MNIST-scale (ideal downsampling) with missing scales.}
\vspace{0.06cm}
\begin{tabular}{@{}lccc}
\specialrule{.15em}{.05em}{.05em}
Models                                 & Acc.$\uparrow$           & Scale-Con.$\uparrow$        & Equi-Err.$\downarrow$        \\ \midrule
\multirow{1}{*}{CNN}                   &0.9842           &  0.7617          &  -                \\

\multirow{1}{*}{Per Res. CNN}          &0.9763          & 0.3594            & -                 \\     
\hline
\multirow{1}{*}{SESN}                  & 0.9892          & \underline{0.8339}           &     -             \\ 
\multirow{1}{*}{DSS}                   & 0.9884          &0.8105            &    -              \\   
\multirow{1}{*}{SI-CovNet}             & 0.9878          &0.6621            &    -               \\   
\multirow{1}{*}{SS-CNN}                &  0.9870        &0.3593             &   -                 \\
\multirow{1}{*}{DISCO}                 & \textbf{0.9914} &  0.5371          & 0.35             \\ \midrule 
\multirow{1}{*}{Fourier CNN}           &0.9820           & 0.1250            & \underline{0.23}               \\
\multirow{1}{*}{Ours}                  &  \underline{0.9888}    &   \textbf{0.9366}         &   \textbf{0.00}       \\
\specialrule{.15em}{.05em}{.05em}
\end{tabular}

\label{tab:mnist_result_ideal_skip}

\end{minipage}
\vspace{-0.1cm}
\end{table}

{\noindent\bf Evaluation metrics.}
To evaluate task performance, we report classification accuracy. Next, we introduce a metric to measure the scale-consistency. Given a sample from the test set, we check whether the cross entropy loss is less than or equal to the classification loss of its down-scaled version. We compute this as a percentage over the dataset and report the scale-consistent rate defined as:
\bea\label{eq:scale_con}
\text{Scale-Con.} = \frac{1}{|\gT|} \sum_{(\rvx,y) \in \gT} \sE_r \Bigg( \mathbf{1}\bigg[ \gL(\rvx,y) \le \gL(\gD_r(\rvx),y)\bigg] \Bigg),
\eea
where $r$ is uniformly sampled over the set of scales for which we want to achieve equivariance and $\mathbf{1}$ denotes the indicator function.

Finally, we quantify the equivariance-error over the final feature map given by a fully trained model on the dataset. The equivariance-error (Equi-Err.) is defined as
\bea 
\text{Equi-Err.} = 
\frac{1}{|\gT||\gS|} \sum_{\rvx \in \gT} \sum_{r \in \gR} \frac{
\norm{g(\gD_r(\rvx)) - \gD_r(g(\rvx))}_2^2
}{\norm{g(\gD_r(\rvx))}_2^2}.
\eea
Here, $\gR$ is the set of all scales over which we enforce equivariance. We report the average equivariance error over the samples of the test set $\gT$. We note that this equivariance-error \textit{differs} from the one reported by~\citet{sosnovik2021disco} where they measured the error for the ``scale-convolution with weights initialized randomly." Contrarily, we measure the equivariance error from \textit{end-to-end} over~\textit{trained} models, which more closely matches how the models are used in practice.

{\noindent\bf Baselines.} Following prior works in scale-equivariant neural networks~\cite{sosnovik2021disco, sosnovik2019scale}
we compare to baselines: DISCO~\cite{sosnovik2021disco},  SI-ConvNet~\cite{kanazawa2014locally}, SS-CNN~\cite{ghosh2019scale}, DSS~\cite{worrall2019deep}, and SESN~\cite{sosnovik2019scale}. For the baseline, we follow the architecture and training scheme provided by~\citet{sosnovik2021disco}. We also prepared three additional baseline models: (a) standard CNN, (b) Per Res, CNN where we train a separate CNN for each resolution in the training set, and (c) Fourier CNN~\cite{li2021fourier} which utilizes Fourier layers.

\subsection{MNIST-scale (Ideal downsampling)}
{\bf Experiment setup.} 
We create the MNIST-scale dataset following the procedure in prior works~\cite{sosnovik2021disco, kanazawa2014locally}. Each image in the original MNIST dataset is randomly downsampled with a factor of $ \sim [\frac{1}{0.3} -1]$, such that every resolution from $8\times 8$ to $28\times 28$ contains an equal number of samples. As the baseline models (except the Fourier CNN) can not handle images of different resolutions, following prior works, lower-resolution images are zero-padded to the original resolution.  We do not need to pad the input for our model and Fourier CNN. We used 10k, 2k, and 50k for training, validation, and test set samples. For this experiment, we enforce equivariance over scales that correspond to the discrete resolutions of $\gR = \{8\time 8, \dots, 28\}$.

{\bf Implementation details.}  
For the baselines and CNN, we follow the implementation, hyper-parameters, and architecture provided in prior works~\cite{sosnovik2021disco, sosnovik2019scale}. For Per Res. CNN, we train a separate CNN for each resolution. Each of these CNNs uses the architecture of baseline CNN. For Fourier CNN, we use the Fourier block introduced in the Fourier Neural operator~\cite{li2021fourier}. Inspired by their design, we use $1\times 1$ complex convolution in the Fourier domain along with the scale-equivariant convolution. We follow the baseline for all training hyper-parameters, except we included a weight decay of $0.01$. 

{\bf Results.} 
In~\tabref{tab:mnist_result_ideal}, we report the accuracy of the MNIST-scale dataset. We observe that our approach achieved zero equivariance error and the highest accuracy. 
While all models achieve similar accuracy, there is a more notable difference in the scale consistency rate. This means that our model properly captures the additional information that comes with increased resolution. 

\begin{table}[t]
\begin{minipage}[c]{0.49\textwidth}

\centering
\small
\caption{MNIST-scale accuracy with different numbers of training samples.
}
\vspace{0.06cm}
\begin{tabular}{lcccc}
\specialrule{.15em}{.05em}{.05em}
Models /~\# Samples     & 5000                  & 2500                      & 1000          \\ \midrule
CNN                       & 0.9432                   & 0.9389                   & 0.8577                   \\
Per Res. CNN              & 0.9118                   & 0.8392                   & 0.5815                   \\
\hline
DISCO                     & \underline{0.9794}                   & \underline{0.9665}                   & \underline{0.9457}                   \\
SESN                      & 0.9638                   & 0.9402                   & 0.9207                   \\
SI-CovNet                 & 0.9641                   & 0.9437                   & 0.9280                    \\
SS-CNN                    & 0.9477                   & 0.9259                   & 0.9176                   \\
DSS                       & 0.9654                   & 0.9401                   & 0.9281                   \\
\hline
Fourier CNN       & 0.9567                   & 0.9419                   & 0.8910                    \\
Ours                      & \textbf{0.9835}                   & \textbf{0.9767}                   & \textbf{0.9606}                   \\
\specialrule{.15em}{.05em}{.05em}
\end{tabular}%

\label{tab:mnist_data_efficiency}

\end{minipage}
\hspace{0.15cm}
\begin{minipage}[c]{0.49\textwidth}
\centering
\small
\caption{The classification accuracy of different models on STL10-scale dataset.
}
\vspace{0.06cm}
\renewcommand{\arraystretch}{1.06}
\begin{tabular}{@{}lccc}
\specialrule{.15em}{.05em}{.05em}
Models                                  & Acc.$\uparrow$              & Scale-Con.$\uparrow$            & Equi-Err.$\downarrow$          \\ \midrule
\multirow{1}{*}{Wide ResNet}            &0.5596               &  0.2916             &0.16             \\
                                    
\multirow{1}{*}{SESN}                   &0.5525             & \underline{0.4166}                & 0.04                \\  
\multirow{1}{*}{DSS}                    &0.5347             &  0.1979               & \underline{0.02}               \\    
\multirow{1}{*}{SI-CovNet}              &0.5588              & 0.2187               & 0.03               \\  
\multirow{1}{*}{SS-CNN}                 & 0.4788            & 0.1979                &1.82                \\
\multirow{1}{*}{DISCO}                  & 0.4768             & 0.3541               &0.06               \\  \midrule
\multirow{1}{*}{Fourier CNN}            & \underline{0.5844}            &  0.2812               & 0.19                \\
\multirow{1}{*}{Ours}                   & \textbf{0.7332}    & \textbf{0.6770}      &   \textbf{0.00}       \\
\specialrule{.15em}{.05em}{.05em}

\end{tabular}

\label{tab:stl_result_ideal}

\end{minipage}
\vspace{-0.2cm}
\end{table}

{\bf Generalization to unseen scales.}
We study the generalization capabilities of the scale-equivariant modes to unseen scales; we train them on a dataset with 10k full resolution $(28\times 28)$ MNIST images and test on 50k samples of MNIST-scale,~\ie, containing different scales. For the baselines, we added random scaling argumentation during training. %
In~\tabref{tab:mnist_result_ideal_skip}, we observe that our model can guarantee zero equivariance error even for the unseen scales and achieves comparable performance to baselines trained with data augmentation.

{\bf Data efficiency.} 
We also conduct experiments studying the data efficiency of the different models. Following the same setup as MNIST-scale, we train the models on limited training examples, 5k, 2.5k, and 1k, of different resolutions and test on 50k samples across all resolutions. In~\tabref{tab:mnist_data_efficiency}, we observe that our model is more data efficient than the baselines. DISCO achieves the second-best performance. We also see that Per Res. CNN suffers the most when trained with fewer data points, as it trains a separate CNN for each scale and does not share parameters across different scales.\vspace{0.15cm}

\begin{table}[t]
\centering
\small

\caption{Ablation on consistency loss.
}

\setlength{\tabcolsep}{3pt}
\begin{tabular}{lcc|cc}
\specialrule{.15em}{.05em}{.05em}
      & \multicolumn{2}{c|}{ w/ {consistency}} & \multicolumn{2}{c}{w/o {consistency}} \\
\# Samples & Acc.$\uparrow$           & Scale-Con.$\uparrow$        & Acc.$\uparrow$             & Scale-Con.$\uparrow$         \\ \midrule
5000  & 0.9835        & 0.9296        & 0.9831          & 0.9150         \\
2500  & 0.9767        & 0.8906        & 0.9755          & 0.8633         \\
1000  & 0.9606        & 0.8183        & 0.9599          & 0.8144         \\
\specialrule{.15em}{.05em}{.05em}
\end{tabular}%

\label{tab:abalation_hing}

\end{table}
{\bf Ablation.} We perform an ablation on the consistency loss in~\equref{eq:hindge} over different training set sizes.  From~\tabref{tab:abalation_hing}, we can observe that the consistency loss improves the accuracy of our model as well as the scale-consistency. This result validates the effectiveness of the proposed consistency loss.

\subsection{STL10-scale (Ideal downsampling)}
{\bf Experiment setup.} Following the same procedure as the MNIST-scale dataset, we create the STL10-scale dataset. Each image of the dataset is randomly scaled with a randomly chosen downsampling factor between $[1 -2]$ such that every resolution from 48 to 97 contains an equal number of samples. We use 7k, 1k, and 5k  samples in our training, validation, and test set. For the baseline models, we again zero-pad the downsampled images to the original size.

{\bf Implementation details.} For the baseline models, we use the Wide ResNet as the CNN baseline following prior work~\cite{sosnovik2021disco, sosnovik2019scale}. For Fourier CNN, we use six Fourier blocks followed by a two-layered MLP. For our model, we use six scale-equivariant Fourier blocks followed by a two-layer MLP. All of the models are trained for $250$ epochs with Adam optimizer with an initial learning rate of $0.01$. The learning rate is reduced by a factor of $0.1$ after every $100$ epoch.
For scalability, we consider achieving equivariance over scales that correspond to the discrete resolutions in the set $\gR = \{48 \le 48+ i \times8 \le 97\}$ $~\forall i \in \{0,1,2, \dots\}$.

{\bf Results.}
In~\tabref{tab:stl_result_ideal}, we observe that our model achieves zero equivariance error with higher accuracy and scale consistency over the baselines. As the baseline models accept a fixed-sized input, the downsampled images are zero-padded following prior work's preprocessing on MNIST-scale. Note, MNIST images have a uniform black background, and zero-padding does not create artifacts. However, for colored images with diverse backgrounds, such as STL-10, any padding scheme to resize the image will cause artifacts. We believe this artifact hurts the performance of baseline models on the STL10-scale dataset. However, it is unclear whether there is a more suitable padding strategy.

\begin{figure*}[t]
    \centering
    \hspace{-0.35cm}
    \subfloat[Learned feature from ideal downsampling.]{
    \hspace{-0.3cm}
    \includegraphics[width=.49\linewidth]{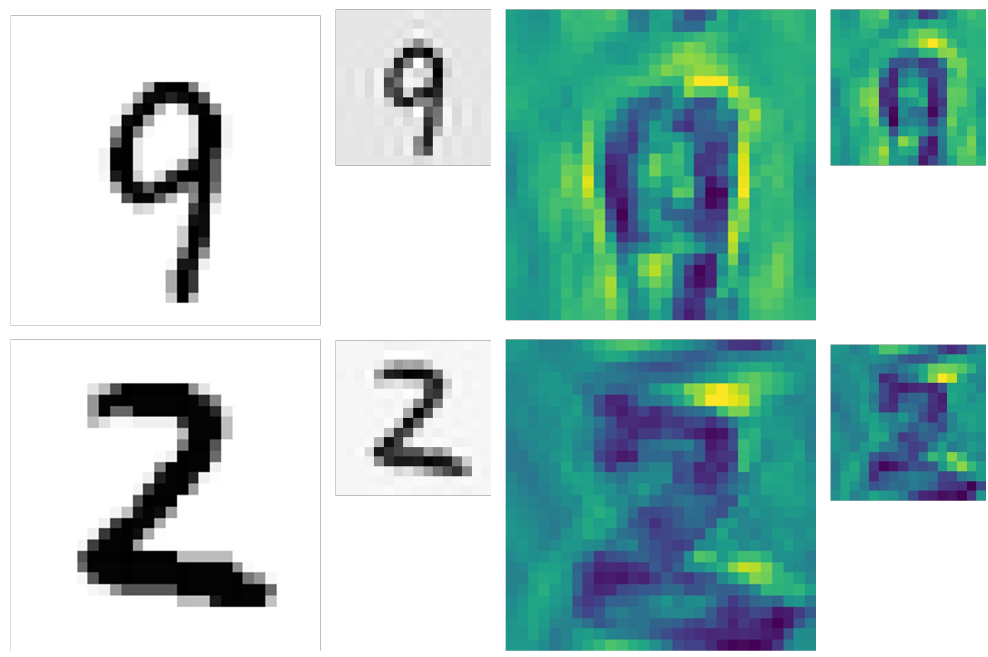}
    \label{fig:feat_ideal}\hspace{0.02\linewidth}
    }
    \subfloat[Learned feature for non-ideal downsampling.]{
    \includegraphics[width=.49\linewidth]{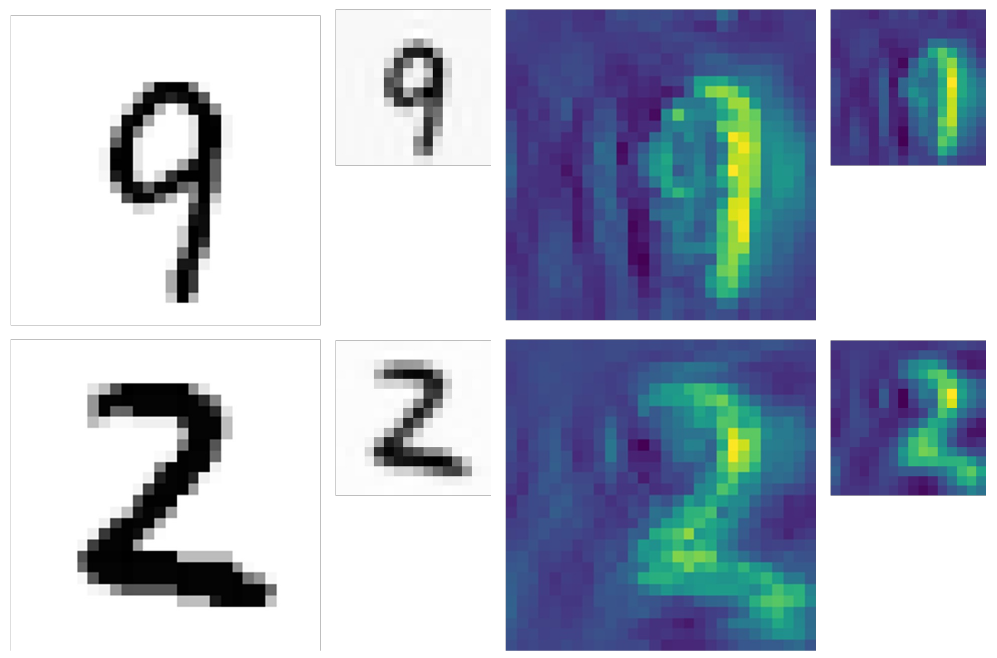}
    \label{fig:feat_non_ideal}
    }
    \caption{Feature visualization for ideal and non-ideal downsampling settings. In both settings, our model seems to learn spatially local features such as digit contour and edges.}
\end{figure*}

 \begin{table}[t]
\centering
\small
\caption{The accuracy of different models on MNIST-scale (non-ideal downsampling).
}
\begin{tabular}{@{}lccc}
\specialrule{.15em}{.05em}{.05em}
Models                                 & Acc.$\uparrow$                    & Scale Con.$\uparrow$           & Equi. Err.$\downarrow$           \\ \midrule
\multirow{1}{*}{CNN}                   &0.9642                  &  0.1033               &  -                     \\
\multirow{1}{*}{Per Res. CNN}          &0.9450                  &0.0742                 &  -                     \\     
\hline
\multirow{1}{*}{SESN}                  & 0.9710                & \underline{0.6666}                  &     -                   \\ 
\multirow{1}{*}{DSS}                   &0.9772                & 0.5716                    &    -                      \\   
\multirow{1}{*}{SI-CovNet}             &0.9694               & 0.4453                      &    -                     \\   
\multirow{1}{*}{SS-CNN}                & 0.9670                &  0.3144                 &   -                     \\
\multirow{1}{*}{DISCO}                 & \underline{0.9830}                  &  0.4500                & 0.63              \\ \midrule 
\multirow{1}{*}{Fourier CNN}            &0.9745                 & 0.1716                & \underline{0.29}                 \\
\multirow{1}{*}{Ours}                  &  \textbf{0.9880}                &   \textbf{0.9760}        &   \textbf{0.05}      \\
\specialrule{.15em}{.05em}{.05em}
\end{tabular}
\label{tab:mnist_result_bicubic}
\end{table}

\subsection{MNIST-scale (Non-ideal downsampling)}%

Ideal interpolation suffers from artifacts known as the ringing effect caused by \textit{Gibbs phenomenon}~\cite{manolakis2011applied}; see the down-scaled image in~\figref{fig:feat_ideal}. In practice, a non-ideal low-pass filter will be used instead. 
Taking this into consideration, we conduct the experiments using a more commonly used anti-aliasing scheme with a Gaussian blur instead of the ideal low-pass filter.\vspace{0.15cm}%

{\bf Experiment details.}
We follow the same experimental setup and training scheme as in MNIST-scale with the ideal downsampling experiment.
The only difference is that 
we use a Gaussian kernel to perform anti-aliasing.\vspace{.15cm}\\
{\bf Results.} From~\tabref{tab:mnist_result_bicubic},
we observe that our model achieves higher classification accuracy and Scale consistency. Importantly, our model achieves lower equivariance error than the baseline despite the gap in the theory of non-ideal downsampling.

\section{Conclusion}
We propose a family of scale-equivariant deep nets that achieve zero equivariance error measured from end to end. We formulate down-scaling in the discrete domain with proper consideration of anti-aliasing. To achieve scale-equivariance, we design novel modules based on Fourier layers, enforcing that the lower frequency content of output does not depend on the higher frequency content of the input. Furthermore, we motivated the scale-consistency property that the performance of higher-resolution input should be better than that of the lower resolution and designed a suitable classifier architecture. Empirically, our approach achieves competitive accuracy on image classification tasks, with improved scale consistency and lower equivariance-error compared to baselines. Similar to other equivariant methodologies, defining consistent scales or group actions to achieve equivalence before constructing the model is crucial. Moreover, a common challenge all equivariant and invariant techniques face is the significant demands on memory and computational resources. In our upcoming research, we plan to enhance our approach by applying it to high-resolution image datasets and dense prediction tasks, such as instance segmentation.
\clearpage 
\bibliographystyle{abbrvnat}
\bibliography{scale_refs.bib}

\begin{thebibliography}{54}
\providecommand{\natexlab}[1]{#1}
\providecommand{\url}[1]{\texttt{#1}}
\expandafter\ifx\csname urlstyle\endcsname\relax
  \providecommand{\doi}[1]{doi: #1}\else
  \providecommand{\doi}{doi: \begingroup \urlstyle{rm}\Url}\fi

\bibitem[Adelson et~al.(1984)Adelson, Anderson, Bergen, Burt, and Ogden]{adelson1984pyramid}
E.~H. Adelson, C.~H. Anderson, J.~R. Bergen, P.~J. Burt, and J.~M. Ogden.
\newblock Pyramid methods in image processing.
\newblock \emph{RCA engineer}, 1984.

\bibitem[Bekkers(2020)]{bekkers2019b}
E.~J. Bekkers.
\newblock {B-spline} {CNNs} on {Lie} groups.
\newblock In \emph{Proc. ICLR}, 2020.

\bibitem[Bronstein et~al.(2017)Bronstein, Bruna, LeCun, Szlam, and Vandergheynst]{bronstein2017geometric}
M.~M. Bronstein, J.~Bruna, Y.~LeCun, A.~Szlam, and P.~Vandergheynst.
\newblock Geometric deep learning: going beyond euclidean data.
\newblock \emph{IEEE SPM}, 2017.

\bibitem[Coates et~al.(2011)Coates, Ng, and Lee]{coates2011analysis}
A.~Coates, A.~Ng, and H.~Lee.
\newblock An analysis of single-layer networks in unsupervised feature learning.
\newblock In \emph{Proc. AISTATS}, 2011.

\bibitem[Cohen and Welling(2016)]{cohen2016group}
T.~Cohen and M.~Welling.
\newblock Group equivariant convolutional networks.
\newblock In \emph{Proc. ICML}, 2016.

\bibitem[de~Haan et~al.(2021)de~Haan, Weiler, Cohen, and Welling]{dehaan2020gauge}
P.~de~Haan, M.~Weiler, T.~Cohen, and M.~Welling.
\newblock Gauge equivariant mesh {CNNs}: Anisotropic convolutions on geometric graphs.
\newblock In \emph{Proc. ICLR}, 2021.

\bibitem[Defferrard et~al.(2016)Defferrard, Bresson, and Vandergheynst]{defferrard2016convolutional}
M.~Defferrard, X.~Bresson, and P.~Vandergheynst.
\newblock Convolutional neural networks on graphs with fast localized spectral filtering.
\newblock In \emph{Proc. NeurIPS}, 2016.

\bibitem[Ghosh and Gupta(2019)]{ghosh2019scale}
R.~Ghosh and A.~K. Gupta.
\newblock Scale steerable filters for locally scale-invariant convolutional neural networks.
\newblock \emph{arXiv preprint arXiv:1906.03861}, 2019.

\bibitem[Grauman and Darrell(2005)]{grauman_2005_pyramid}
K.~Grauman and T.~Darrell.
\newblock The pyramid match kernel: discriminative classification with sets of image features.
\newblock In \emph{Proc. ICCV}, 2005.

\bibitem[Hartford et~al.(2018)Hartford, Graham, Leyton-Brown, and Ravanbakhsh]{hartford2018deep}
J.~Hartford, D.~Graham, K.~Leyton-Brown, and S.~Ravanbakhsh.
\newblock Deep models of interactions across sets.
\newblock In \emph{Proc. ICML}, 2018.

\bibitem[He et~al.(2015)He, Zhang, Ren, and Sun]{he2015spatial}
K.~He, X.~Zhang, S.~Ren, and J.~Sun.
\newblock Spatial pyramid pooling in deep convolutional networks for visual recognition.
\newblock \emph{IEEE TPAMI}, 2015.

\bibitem[He et~al.(2021)He, Chen, Dong, Wang, Lin, et~al.]{he2021efficient}
L.~He, Y.~Chen, Y.~Dong, Y.~Wang, Z.~Lin, et~al.
\newblock Efficient equivariant network.
\newblock In \emph{Proc. NeurIPS}, 2021.

\bibitem[Kanazawa et~al.(2014)Kanazawa, Sharma, and Jacobs]{kanazawa2014locally}
A.~Kanazawa, A.~Sharma, and D.~Jacobs.
\newblock Locally scale-invariant convolutional neural networks.
\newblock \emph{arXiv preprint arXiv:1412.5104}, 2014.

\bibitem[Karras et~al.(2021)Karras, Aittala, Laine, H{\"a}rk{\"o}nen, Hellsten, Lehtinen, and Aila]{karras2021alias}
T.~Karras, M.~Aittala, S.~Laine, E.~H{\"a}rk{\"o}nen, J.~Hellsten, J.~Lehtinen, and T.~Aila.
\newblock Alias-free generative adversarial networks.
\newblock In \emph{Proc. NeurIPS}, 2021.

\bibitem[Kipf and Welling(2017)]{kipf2017semi}
T.~N. Kipf and M.~Welling.
\newblock Semi-supervised classification with graph convolutional networks.
\newblock In \emph{Proc. ICLR}, 2017.

\bibitem[Kondor et~al.(2018)Kondor, Lin, and Trivedi]{kondor2018clebsch}
R.~Kondor, Z.~Lin, and S.~Trivedi.
\newblock {Clebsch\textendash Gordan Nets}: a fully {Fourier} space spherical convolutional neural network.
\newblock In \emph{Proc. NeurIPS}, 2018.

\bibitem[Lazebnik et~al.(2006)Lazebnik, Schmid, and Ponce]{lazebnik2006beyond}
S.~Lazebnik, C.~Schmid, and J.~Ponce.
\newblock Beyond bags of features: Spatial pyramid matching for recognizing natural scene categories.
\newblock In \emph{Proc. CVPR}, 2006.

\bibitem[Li et~al.(2021)Li, Kovachki, Azizzadenesheli, liu, Bhattacharya, Stuart, and Anandkumar]{li2021fourier}
Z.~Li, N.~B. Kovachki, K.~Azizzadenesheli, B.~liu, K.~Bhattacharya, A.~Stuart, and A.~Anandkumar.
\newblock Fourier neural operator for parametric partial differential equations.
\newblock In \emph{Proc. ICLR}, 2021.

\bibitem[Liu et~al.(2020)Liu, Yeh, and Schwing]{liu2020pic}
I.-J. Liu, R.~A. Yeh, and A.~G. Schwing.
\newblock Pic: permutation invariant critic for multi-agent deep reinforcement learning.
\newblock In \emph{Proc. CORL}, 2020.

\bibitem[Liu et~al.(2021)Liu, Ren, Yeh, and Schwing]{liu2021semantic}
I.-J. Liu, Z.~Ren, R.~A. Yeh, and A.~G. Schwing.
\newblock Semantic tracklets: An object-centric representation for visual multi-agent reinforcement learning.
\newblock In \emph{Proc. IROS}, 2021.

\bibitem[Lowe(1999)]{lowe1999object}
D.~G. Lowe.
\newblock Object recognition from local scale-invariant features.
\newblock In \emph{Proc. ICCV}, 1999.

\bibitem[Lowe(2004)]{lowe2004distinctive}
D.~G. Lowe.
\newblock Distinctive image features from scale-invariant keypoints.
\newblock \emph{IJCV}, 2004.

\bibitem[Manolakis and Ingle(2011)]{manolakis2011applied}
D.~G. Manolakis and V.~K. Ingle.
\newblock \emph{Applied digital signal processing: theory and practice}.
\newblock Cambridge university press, 2011.

\bibitem[Mao et~al.(2023)Mao, Liu, Liu, Li, Shen, and Wang]{xint2023freqsel}
X.~Mao, Y.~Liu, F.~Liu, Q.~Li, W.~Shen, and Y.~Wang.
\newblock Intriguing findings of frequency selection for image deblurring.
\newblock In \emph{Proc. AAAI}, 2023.

\bibitem[Maron et~al.(2019)Maron, Ben-Hamu, Shamir, and Lipman]{maron2018invariant}
H.~Maron, H.~Ben-Hamu, N.~Shamir, and Y.~Lipman.
\newblock Invariant and equivariant graph networks.
\newblock In \emph{Proc. ICLR}, 2019.

\bibitem[Maron et~al.(2020)Maron, Litany, Chechik, and Fetaya]{maron2020learning}
H.~Maron, O.~Litany, G.~Chechik, and E.~Fetaya.
\newblock On learning sets of symmetric elements.
\newblock In \emph{Proc. ICML}, 2020.

\bibitem[Mathieu et~al.(2013)Mathieu, Henaff, and LeCun]{mathieu2013fast}
M.~Mathieu, M.~Henaff, and Y.~LeCun.
\newblock Fast training of convolutional networks through {FFTs}.
\newblock \emph{arXiv preprint arXiv:1312.5851}, 2013.

\bibitem[Morris et~al.(2022)Morris, Rattan, Kiefer, and Ravanbakhsh]{morris22a}
C.~Morris, G.~Rattan, S.~Kiefer, and S.~Ravanbakhsh.
\newblock {S}peq{N}ets: Sparsity-aware permutation-equivariant graph networks.
\newblock In \emph{Proc. ICML}, 2022.

\bibitem[Nguyen et~al.(2022)Nguyen, Goel, Gu, Downs, Shah, Dao, Baccus, and R{\'e}]{nguyen2022s4nd}
E.~Nguyen, K.~Goel, A.~Gu, G.~Downs, P.~Shah, T.~Dao, S.~Baccus, and C.~R{\'e}.
\newblock S4nd: Modeling images and videos as multidimensional signals with state spaces.
\newblock In \emph{Proc. NeurIPS}, 2022.

\bibitem[Nyquist(1928)]{nyquist1928certain}
H.~Nyquist.
\newblock Certain topics in telegraph transmission theory.
\newblock \emph{Transactions of the American Institute of Electrical Engineers}, 1928.

\bibitem[Pratt et~al.(2017)Pratt, Williams, Coenen, and Zheng]{pratt2017fcnn}
H.~Pratt, B.~Williams, F.~Coenen, and Y.~Zheng.
\newblock {FCNN}: Fourier convolutional neural networks.
\newblock In \emph{ECML PKDD}, 2017.

\bibitem[Qi et~al.(2017)Qi, Su, Mo, and Guibas]{qi2017pointnet}
C.~R. Qi, H.~Su, K.~Mo, and L.~J. Guibas.
\newblock {PointNet}: Deep learning on point sets for {3D} classification and segmentation.
\newblock In \emph{Proc. CVPR}, 2017.

\bibitem[Ravanbakhsh et~al.(2017{\natexlab{a}})Ravanbakhsh, Schneider, and P{\'o}czos]{ravanbakhsh2017equivariance}
S.~Ravanbakhsh, J.~Schneider, and B.~P{\'o}czos.
\newblock Equivariance through parameter-sharing.
\newblock In \emph{Proc. ICML}, 2017{\natexlab{a}}.

\bibitem[Ravanbakhsh et~al.(2017{\natexlab{b}})Ravanbakhsh, Schneider, and Poczos]{ravanbakhsh_sets}
S.~Ravanbakhsh, J.~Schneider, and B.~Poczos.
\newblock Deep learning with sets and point clouds.
\newblock In \emph{Proc. ICLR workshop}, 2017{\natexlab{b}}.

\bibitem[Rojas-Gomez et~al.(2022)Rojas-Gomez, Lim, Schwing, Do, and Yeh]{rojas2022learnable}
R.~A. Rojas-Gomez, T.-Y. Lim, A.~Schwing, M.~Do, and R.~A. Yeh.
\newblock Learnable polyphase sampling for shift invariant and equivariant convolutional networks.
\newblock In \emph{Proc. NeurIPS}, 2022.

\bibitem[Rojas-Gomez et~al.(2023)Rojas-Gomez, Lim, Do, and Yeh]{rojas2023making}
R.~A. Rojas-Gomez, T.-Y. Lim, M.~N. Do, and R.~A. Yeh.
\newblock Making vision transformers truly shift-equivariant.
\newblock \emph{arXiv preprint arXiv:2305.16316}, 2023.

\bibitem[Romero et~al.(2020)Romero, Bekkers, Tomczak, and Hoogendoorn]{romero2020attentive}
D.~Romero, E.~Bekkers, J.~Tomczak, and M.~Hoogendoorn.
\newblock Attentive group equivariant convolutional networks.
\newblock In \emph{Proc. ICML}, 2020.

\bibitem[Shakerinava and Ravanbakhsh(2021)]{shakerinava21a}
M.~Shakerinava and S.~Ravanbakhsh.
\newblock Equivariant networks for pixelized spheres.
\newblock In \emph{Proc. ICML}, 2021.

\bibitem[Shuman et~al.(2013)Shuman, Narang, Frossard, Ortega, and Vandergheynst]{shuman2013emerging}
D.~I. Shuman, S.~K. Narang, P.~Frossard, A.~Ortega, and P.~Vandergheynst.
\newblock The emerging field of signal processing on graphs: Extending high-dimensional data analysis to networks and other irregular domains.
\newblock \emph{IEEE SPM}, 2013.

\bibitem[Sohn and Lee(2012)]{sohn2012learning}
K.~Sohn and H.~Lee.
\newblock Learning invariant representations with local transformations.
\newblock In \emph{Proc. ICML}, 2012.

\bibitem[Sosnovik et~al.(2020)Sosnovik, Szmaja, and Smeulders]{sosnovik2019scale}
I.~Sosnovik, M.~Szmaja, and A.~Smeulders.
\newblock Scale-equivariant steerable networks.
\newblock In \emph{Proc. ICLR}, 2020.

\bibitem[Sosnovik et~al.(2021)Sosnovik, Moskalev, and Smeulders]{sosnovik2021disco}
I.~Sosnovik, A.~Moskalev, and A.~Smeulders.
\newblock {DISCO}: accurate discrete scale convolutions.
\newblock In \emph{Proc. BMVC}, 2021.

\bibitem[Suvorov et~al.(2022)Suvorov, Logacheva, Mashikhin, Remizova, Ashukha, Silvestrov, Kong, Goka, Park, and Lempitsky]{suvorov2022resolution}
R.~Suvorov, E.~Logacheva, A.~Mashikhin, A.~Remizova, A.~Ashukha, A.~Silvestrov, N.~Kong, H.~Goka, K.~Park, and V.~Lempitsky.
\newblock Resolution-robust large mask inpainting with {Fourier} convolutions.
\newblock In \emph{Proc. WACV}, 2022.

\bibitem[Venkataraman et~al.(2020)Venkataraman, Balasubramanian, and Sarma]{venkataraman2019building}
S.~R. Venkataraman, S.~Balasubramanian, and R.~R. Sarma.
\newblock Building deep equivariant capsule networks.
\newblock In \emph{Proc. ICLR}, 2020.

\bibitem[Weiler and Cesa(2019)]{weiler2019general}
M.~Weiler and G.~Cesa.
\newblock General {E(2)}-equivariant steerable {CNNs}.
\newblock In \emph{Proc. NeurIPS}, 2019.

\bibitem[Worrall and Welling(2019)]{worrall2019deep}
D.~Worrall and M.~Welling.
\newblock Deep scale-spaces: Equivariance over scale.
\newblock In \emph{Proc. NeurIPS}, 2019.

\bibitem[Xu et~al.(2021)Xu, Kim, Rainforth, and Teh]{xu2021group}
J.~Xu, H.~Kim, T.~Rainforth, and Y.~Teh.
\newblock Group equivariant subsampling.
\newblock In \emph{Proc. NeurIPS}, 2021.

\bibitem[Yeh et~al.(2019{\natexlab{a}})Yeh, Hu, and Schwing]{yeh2019chirality}
R.~A. Yeh, Y.-T. Hu, and A.~Schwing.
\newblock Chirality nets for human pose regression.
\newblock In \emph{Proc. NeurIPS}, 2019{\natexlab{a}}.

\bibitem[Yeh et~al.(2019{\natexlab{b}})Yeh, Schwing, Huang, and Murphy]{yeh2019diverse}
R.~A. Yeh, A.~G. Schwing, J.~Huang, and K.~Murphy.
\newblock Diverse generation for multi-agent sports games.
\newblock In \emph{Proc. CVPR}, 2019{\natexlab{b}}.

\bibitem[Yeh et~al.(2022)Yeh, Hu, Hasegawa-Johnson, and Schwing]{yeh2022equivariance}
R.~A. Yeh, Y.-T. Hu, M.~Hasegawa-Johnson, and A.~Schwing.
\newblock Equivariance discovery by learned parameter-sharing.
\newblock In \emph{Proc. AISTATS}, 2022.

\bibitem[Zaheer et~al.(2017)Zaheer, Kottur, Ravanbakhsh, Poczos, Salakhutdinov, and Smola]{zaheer2017deep}
M.~Zaheer, S.~Kottur, S.~Ravanbakhsh, B.~Poczos, R.~R. Salakhutdinov, and A.~J. Smola.
\newblock Deep sets.
\newblock In \emph{Proc. NeurIPS}, 2017.

\bibitem[Zhang(2019)]{zhang2019making}
R.~Zhang.
\newblock Making convolutional networks shift-invariant again.
\newblock In \emph{Proc. ICML}, 2019.

\bibitem[Zhao et~al.(2017)Zhao, Shi, Qi, Wang, and Jia]{zhao2017pyramid}
H.~Zhao, J.~Shi, X.~Qi, X.~Wang, and J.~Jia.
\newblock Pyramid scene parsing network.
\newblock In \emph{Proc. CVPR}, 2017.

\bibitem[Zhu et~al.(2022)Zhu, Qiu, Calderbank, Sapiro, and Cheng]{zhu2022scaling}
W.~Zhu, Q.~Qiu, R.~Calderbank, G.~Sapiro, and X.~Cheng.
\newblock Scaling-translation-equivariant networks with decomposed convolutional filters.
\newblock \emph{JMLR}, 2022.

\end{thebibliography}

\newpage
\appendix
\onecolumn
\setcounter{section}{0}
\renewcommand{\thesection}{A\arabic{section}}
\renewcommand{\thetable}{A\arabic{table}}
\setcounter{table}{0}
\setcounter{figure}{0}
\renewcommand{\thetable}{A\arabic{table}}
\renewcommand\thefigure{A\arabic{figure}}
\renewcommand{\theHtable}{A.Tab.\arabic{table}}%
\renewcommand{\theHfigure}{A.Abb.\arabic{figure}}%
\renewcommand\theequation{A\arabic{equation}}
\renewcommand{\theHequation}{A.Abb.\arabic{equation}}%

{\bf \LARGE Appendix:}
\vspace{0.1cm}

The appendix is organized as follows:
\vspace{-0.1cm}
\begin{itemize}
\item In~\secref{supp_sec:claim}, we provide a generalization of Claim~\ref{cla:main_claim} and its the complete proof.
\item In~\secref{supp_sec:proof}, we provide the complete proof for~\equref{eq:local_kernel}.
\item In~\secref{supp_sec:result}, we provide additional ablations and experimental results.
\item In~\secref{sec:supp_implt}, we provide additional implementation details.
\end{itemize}

\section{Generalization of  Claim~\ref{cla:main_claim}}\label{supp_sec:claim}
In the main paper, we provide proof where we assume that the input and output of the deep net are of the same size. Here, we provide a generalization without such an assumption.

\begin{mdframed}[style=MyFrame,align=center]
\begin{claim}
\vspace{.2em}
Let $g$ denote a deep net such that $\rvy = g(\rvx)$ where $\rvx \in \R^m$, $\rvy \in \R^{\frac{m}{a}}$, and $\frac{m}{a}$ is an integer $~\forall m \in \mathbb{N}$ and $a\ge 1$.
If this deep net $\tt g$ can be equivalently represented as a set of functions $\tilde{G}_k: \Co^{2 ak  +1} \rightarrow \Co$ such that
\bea
\rmY[k]= \tilde{G}_k (\rmX[- ak: ak]
)~~\forall k 
\eea
then $g$ is scale-equivariant as defined in~\equref{eq:scale_eq} for all scales that creates resolutions at a multiple of $a$ after scaling the input vector $\rvx$.
\label{cla:_claim}
\end{claim}
\end{mdframed}

\begin{proof}
We denote the deep net's input and output as $\rvx$ and $\rvy$ with corresponding DFT $\rmX$ and $\rmY$.
We further denote that the deep net's down-scaled input and output as $\rvx' = \gD_R(\rvx)$ and $\rvy'=g(\rvx')$ with corresponding DFT $\rmX'$ and $\rmY'$. 

Given a deep net $g: \R^n \rightarrow \R^{\frac{n}{a}} ~~ \forall n \in \{1,2, \dots N\}$ is a deep net that satisfies Claim~\ref{cla:_claim} then
\bea
&\rmY[k] &= \tilde{G}_k(\rmX[-ak:ak]) ~~ \forall k \le \frac{N}{R}\\
&         &=\tilde{G}_k(\rmX'[-ak:ak]) = \rmY'[k] \text{~~Following the property of } \gD_R \text{ in~\equref{eq:scale_opt_fourier_rel}} %
\eea
Therefore, $\forall k \le \frac{N}{R}~~ \rmY[k] = \rmY'[k]$. By the definition of ideal downsampling $\rmY' = \gD_R(\rmY)$, ${g}(\gD_R(\rvx)) = \gD_R({g}(\rvx))$ concluding that $g$ is scale-equivariant. Note that $\frac{n}{a}$ needs to be an integer for $\R^\frac{n}{a}$ to be a valid vector, \ie,  the resolution of $\rvx$ needs to be a multiple of $a$.
\end{proof}

\section{Proof for Spatially Localized Fourier Layer in~\equref{eq:local_kernel}}\label{supp_sec:proof}
In the main paper, we introduced the parameterization for a spatially localized Fourier layer: 
\be
 \rmK[p] = \frac{1}{d} \sum_{m = -\frac{l}{2}}^{\frac{l}{2}} \big( \rmK^l[m] \sum_{n=0}^{l-1}e^{-2 \pi jn( \frac{p}{d}- \frac{m}{l})} \big).\tag{\ref{eq:local_kernel}}
\ee
We now provide the derivation.

\begin{proof}
From definition of DFT, $\rvk$ can be written as
\bea
&\rmK[p]&= \frac{1}{d} \sum_{n = 0}^{d-1} \rvk[n] e^{-2j\pi \frac{np}{d}} =\frac{1}{d} \sum_{n = 0}^{d-1} \rvk^l[n] e^{-2j\pi \frac{np}{d}}~~(\text{Ignorig the } 0 \text{ elements})\\
&&=\frac{1}{d} \sum_{n = 0}^{d-1}  \sum_{m = -\frac{l}{2}}^{\frac{l}{2}}\rmK^l[m] e^{2j\frac{mn}{l}} e^{-2j\pi \frac{np}{d}}~~(\text{Using DFT of kernel } \rvk^l)\\
&&=\frac{1}{d} \sum_{m = -\frac{l}{2}}^{\frac{l}{2}} \big( \rmK^l[m] \sum_{n=0}^{l-1}e^{-2 \pi jn( \frac{p}{d}- \frac{m}{l})} \big)~~(\text{Exchanging the Sum)}
\eea
Finally, the Geometric series $\sum_{n=0}^{l-1}e^{-2 \pi jnq}$ can be expressed as 
\bea 
\sum_{n=0}^{l-1}e^{-2 \pi jnq} = e^{-j q \frac{l-1}{2}}\frac{\sin(l \frac{q}{2})}{\sin(\frac{q}{2})} \text{ where } \lim_{q \rightarrow 0} \frac{\sin(l\frac{q}{2})}{\sin(\frac{q}{2})} = l.
\eea
\end{proof}

\section{Additional Results}\label{supp_sec:result}

{\bf Additional ablations.}
We conduct additional ablation studies for the proposed spatially local Fourier Layer and scale-equivariant ReLU and report the results in~\tabref{tab:supp_ablation}. We observe that there is a drop in accuracy of 0.5\% when not using spatially local Fourier layers and that the equivariance error greatly increases if we do not use our proposed scale-equivariant ReLU. 

\begin{table}[ht!]
\centering
\small
\caption{Ablation of Spatially localized Fourier filter and Scale-equivariant non-linearity }
\label{tab:supp_ablation}
\begin{tabular}{lccc}
\specialrule{.15em}{.05em}{.05em}
Models & Acc.$\uparrow$              & Scale-Con.$\uparrow$            & Equi-Err.$\downarrow$          \\ \midrule
Ours                      & 0.9889 & 0.9716    & 0.00      \\
Ours w/o Local Filter     & 0.9835 & 0.9628    & 0.00      \\
Ours w/o Scale-equi. ReLU & 0.9897 & 0.9492    & 7.32      \\
Fourier CNN               & 0.9713 & 0.2421    & 0.28     \\
\specialrule{.15em}{.05em}{.05em}
\end{tabular}

\end{table}

{\bf Ablation on baseline's preprocessing (Zero-padding vs. ideal upsampling).}
In the main paper, all the baselines use zero-padding to pre-process images at different resolutions to the same size following prior works. However, we suspect that zero-padding on color images may hurt model performance. In~\tabref{tab:supp_stl_ideal_up}, we provide additional experimental results by performing an ideal upsampling for the baselines. We observe that there are improvements in accuracy for the baseline models. However, our proposed model still achieves the best accuracy with the lowest equivariance-error. 

\begin{table}[ht!]
\centering
\small
\caption{The classification accuracy of different models on STL10-scale dataset with \textbf{ideal} downsampling. For baseline models, images at different scales are resized via an \textbf{ideal upsampling operation.}}
\label{tab:supp_stl_ideal_up}
\begin{tabular}{@{}lccc}
\specialrule{.15em}{.05em}{.05em}
Models                                  & Acc.$\uparrow$              & Scale-Con.$\uparrow$            & Equi-Err.$\downarrow$          \\ \midrule
\multirow{1}{*}{Wide ResNet}            &0.6040               &  0.4791             & 0.20        \\
                                    
\multirow{1}{*}{SESN}                   &0.6428             & 0.5629                & 0.08                \\  
\multirow{1}{*}{DSS}                    &0.6131             &  0.6562              & \underline{0.02}               \\    
\multirow{1}{*}{SI-CovNet}              &0.6722              & 0.3854               & 0.03              \\  
\multirow{1}{*}{SS-CNN}                 & 0.3246            & 0.5833                & 0.04                \\
\multirow{1}{*}{DISCO}                  & 0.5670             & 0.4791               & 0.05              \\  \midrule
\multirow{1}{*}{Fourier CNN}            & 0.5844            &  0.2812               & 0.19                \\
\multirow{1}{*}{Ours}                   & \textbf{0.7332}    & \textbf{0.6770}      &   \textbf{0.00}       \\
\specialrule{.15em}{.05em}{.05em}
\end{tabular}
\end{table}

In~\tabref{tab:supp:stl_ideal_pad} and~\tabref{tab:supp:stl_ideal_up}, we report the same ablation of zero-padding vs. ideal upsampling for the baselines. For this non-ideal downsampling setting, we observe that DSS has the lowest equivariance error, with ours achieving the second best. Our model achieves the highest accuracy out of all the models.

\begin{table}[h]
\begin{minipage}[c]{0.49\textwidth}
\vspace{0.06cm}
\centering
\small
\caption{The classification accuracy of different models on STL10-scale dataset with {\bf non-ideal} downsampling. For baseline models, images at different scales are resized via {\bf zero-padding}.}
\label{tab:supp:stl_ideal_pad}
\begin{tabular}{@{}lccc}
\specialrule{.15em}{.05em}{.05em}
Models                                  & Acc.$\uparrow$              & Scale-Con.$\uparrow$            & Equi-Err.$\downarrow$          \\ \midrule
\multirow{1}{*}{Wide ResNet}            &0.4456               &  0.3229             &1.08           \\
                                    
\multirow{1}{*}{SESN}                   &0.5155             & 0.4687                & 0.07               \\  
\multirow{1}{*}{DSS}                    &0.4756             &  0.3645              & \textbf{0.03}               \\    
\multirow{1}{*}{SI-CovNet}              &0.5234              & 0.3958               & 0.07              \\  
\multirow{1}{*}{SS-CNN}                 & 0.3418            & 0.2187                & 1.72              \\
\multirow{1}{*}{DISCO}                  & 0.5125             & 0.4479               & 0.12              \\  \midrule
\multirow{1}{*}{Fourier CNN}            & \underline{0.5357}            &  0.5312               & 0.20                \\
\multirow{1}{*}{Ours}                   & \textbf{0.7262}    & \textbf{0.5624}      &   0.06       \\
\specialrule{.15em}{.05em}{.05em}
\end{tabular}

\end{minipage}
\hspace{0.45cm}
\begin{minipage}[c]{0.49\textwidth}
\vspace{0.06cm}
\centering
\small
\caption{The classification accuracy of different models on STL10-scale dataset with {\bf non-ideal} downsampling. For baseline models, images at different scales are resized via an \bf{ideal upsampling} operation.}
\label{tab:supp:stl_ideal_up}
\begin{tabular}{@{}lccc}
\specialrule{.15em}{.05em}{.05em}
Models                                  & Acc.$\uparrow$              & Scale-Con.$\uparrow$            & Equi-Err.$\downarrow$          \\ \midrule
\multirow{1}{*}{Wide ResNet}            &0.5952              &  0.4791             &0.75           \\
                                    
\multirow{1}{*}{SESN}                   &0.6312             & 0.5208               & 0.08              \\  
\multirow{1}{*}{DSS}                    &0.6126             &  0.5208              & {\bf 0.03}              \\    
\multirow{1}{*}{SI-CovNet}              &\underline{0.6337}              & 0.4062               & 0.04             \\  
\multirow{1}{*}{SS-CNN}                 & 0.4855            & 0.3854                &  0.05            \\
\multirow{1}{*}{DISCO}                  & 0.5191             & 0.4687               & 0.04              \\  \midrule
\multirow{1}{*}{Fourier CNN}            & 0.5357            &  0.5312               & 0.20                \\
\multirow{1}{*}{Ours}                   & \textbf{0.7262}    & \textbf{0.5624}      &   0.06       \\
\specialrule{.15em}{.05em}{.05em}
\end{tabular}

\end{minipage}
\end{table}

{\bf Ablations of the effect of different types of padding on baselines}\\
In Table \ref{tab:padding_acc}, we present the classification accuracy of various baselines on the STL1-scale dataset, employing ideal-downsampling and padding techniques, including Replicate, Circular, and Reflect.
\begin{table}[ht!]
\centering
\small

\caption{The classification accuracy of different Scale-equivariant baseline models on STL10-scale dataset with different padding strategies.}
\label{tab:padding_acc}
\begin{tabular}{@{}lccc}
\specialrule{.15em}{.05em}{.05em}
Models                                  &Replicate	           &Circular	        &Reflect         \\ \midrule                    \multirow{1}{*}{SESN}                   &0.64	               &0.65	            &0.50               \\  
\multirow{1}{*}{DSS}                    &0.61	               &0.48	            &0.49               \\    
\multirow{1}{*}{SI-CovNet}              &0.54	               &0.63	            &0.63           \\  
\multirow{1}{*}{SS-CNN}                 &0.47	               &0.50	            &0.50                \\
\multirow{1}{*}{DISCO}                  &0.60	               &0.52	            &0.44             \\  \midrule
\specialrule{.15em}{.05em}{.05em}
\end{tabular}
\end{table}

\section{Additional implementation details}\label{sec:supp_implt}

Please refer to \textit{our attached code} in the supplementary materials for more implementation details. Below, we briefly describe the model architectures. 
For the MNIST-scale dataset, the spatially localized Fourier layers use a locality size of  $7\times7$ and $11 \times 11$. For the SLT10-scale dataset, we use a special locality of size $5 \times 5$ for all the localized Fourier layers. In both MNIST-scale and STL10-scale experiments, we use a 2D max-pooling layer before passing the flattened scale-equivariant spatial feature to the MLP. For the scale-equivariant non-linearity, we also apply instance normalization before applying point-wise non-linearity. All the models are trained on a single NVIDIA RTX 3090.

\end{document}